\documentclass[10pt,twocolumn,letterpaper]{article}

\usepackage{iccv}
\usepackage{times}
\usepackage{epsfig}
\usepackage{graphicx}
\usepackage{amsmath}
\usepackage{amssymb}
\usepackage{booktabs} 
\usepackage[english]{babel}
\usepackage[utf8]{inputenc}

\usepackage[noend]{algpseudocode}
\usepackage[linesnumbered,ruled]{algorithm2e}
\usepackage{dsfont}

\usepackage[pagebackref=true,breaklinks=true,letterpaper=true,colorlinks,bookmarks=false]{hyperref}

\iccvfinalcopy 

\def\iccvPaperID{664} 

\ificcvfinal\fi
\begin{document}

\title{Temporal Dynamic Graph LSTM for Action-driven Video Object Detection}
\author{Yuan Yuan$ ^1 $~~~~Xiaodan Liang$ ^2 $~~~~Xiaolong Wang$ ^2 $~~~~Dit-Yan Yeung$ ^1 $~~~~Abhinav Gupta$ ^2 $\\
$ ^1 $The Hong Kong University of Science and Technology~~~$ ^2 $ Carneige Mellon University\\
{\tt\footnotesize yyuanad@ust.hk, xiaodan1@cs.cmu.edu, xiaolonw@cs.cmu.edu, dyyeung@cse.ust.hk, abhinavg@cs.cmu.edu}}

\maketitle

\begin{abstract}
In this paper, we investigate a weakly-supervised object detection framework. Most existing frameworks focus on using static images to learn object detectors. However, these detectors often fail to generalize to videos because of the existing domain shift. Therefore, we investigate learning these detectors directly from boring videos of daily activities. Instead of using bounding boxes, we explore the use of action descriptions as supervision since they are relatively easy to gather. A common issue, however, is that objects of interest that are not involved in human actions are often absent in global action descriptions known as ``missing label''. To tackle this problem, we propose a novel temporal dynamic graph Long Short-Term Memory network (TD-Graph LSTM). TD-Graph LSTM enables global temporal reasoning by constructing a dynamic graph that is based on temporal correlations of object proposals and spans the entire video. The missing label issue for each individual frame can thus be significantly alleviated by transferring knowledge across correlated objects proposals in the whole video. Extensive evaluations on a large-scale daily-life action dataset (i.e., {Charades}) demonstrates the superiority of our proposed method. We also release object bounding-box annotations for more than 5,000 frames in Charades. We believe this annotated data can also benefit other research on video-based object recognition in the future.

\end{abstract}

\section{Introduction}
\begin{figure*}[!tp]
  	\begin{center}
  		\includegraphics[width=0.85\linewidth]{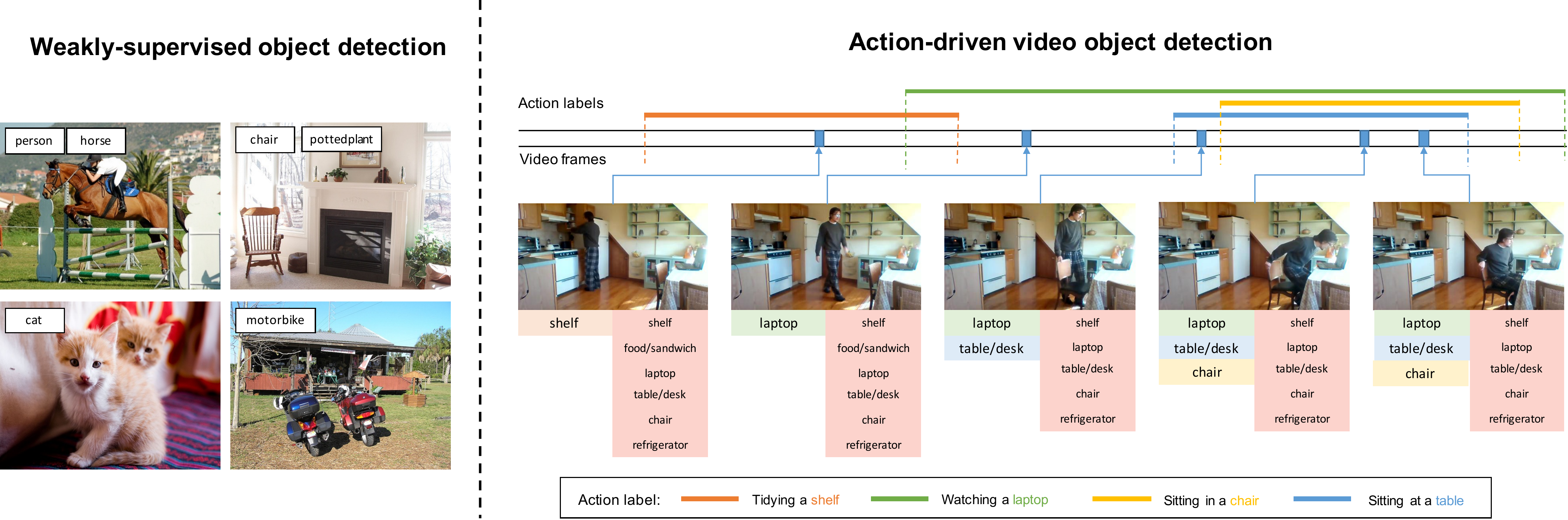}
  		\caption{{\textbf{(Left)} shows the traditional weakly-supervised object detection setting. Each training image has an accurate image-level annotation about object categories. \textbf{(Right)} shows our action-driven weakly-supervised video detection setting. Video-level action labels are provided for each video, indicating what and when (the start and end) the action happened in the video. For each frame, the object categories in its left-below are the participating objects in the action label, while those in its right-below are all objects appearing in the frame.}}
  		\label{fig:problem}
  		\vspace{-6mm}
  	\end{center}
  \end{figure*}

With the recent success of data-driven approaches in recognition, there has been a growing interest in scaling up object detection systems~\cite{Sun2017}. However, unlike classification,  exhaustively annotating object instances with diverse classes and 
bounding boxes is hardly scalable. Therefore, there has been a surge in exploring in unsupervised and weakly-supervised approaches for object detection. However, fully unsupervised approaches \cite{schulter2013unsupervised,Suha15} without any annotations currently give considerably inferior performance on similar tasks, while conventional weakly-supervised methods~\cite{bilen2016weakly,kantorov2016contextlocnet,wang2016video} use static images to learn the detectors. These object detectors, however, fail to generalize to videos due to shift in domain. One alternative is to use these weakly-supervised approaches but using video frames themselves. However, current approaches rely heavily on the accuracy of image-level labels and are vulnerable to missing labels (as shown in Figure~\ref{fig:problem}). Can we design a learning framework that is robust to these missing labels ?

In this paper, we explore a novel slightly-supervised video object detection pipeline that uses human action labels as supervision for object detection. As illustrated in Figure~\ref{fig:problem}, the coarse human action labels spanning multiple frames (e.g., \textit{watching a laptop} or \textit{sitting in a chair}) help indicate the presence of participating object instances (e.g., \textit{laptop} and \textit{chair}). Compared to prior works, our investigated setting has two major merits: 1) the textual action descriptions for videos are much cheaper to collect, e.g., through text tags, search queries and action recognition datasets~\cite{SigurdssonVWFLG16,HeilbronEGN15,UCF101}; and 2) the intrinsic temporal coherence in video domain provides more cues to facilitate the recognition of each object instance and help overcome the missing label problem.

Action-driven supervision for object detection is much more challenging since it can only access object labels for some specific frames, while a considerable number of uninvolved object labels are unknown. As shown in the right column of Figure~\ref{fig:problem}, four action categories are labeled for different periods in the given video. In each period, the action label (e.g., \emph{tidying a shelf}) only points out the \textit{shelf} category and misses the rest of the categories such as \textit{laptop}, \textit{table}, \textit{chair} and \textit{refrigerator}. On the other hand, the missed categories (e.g., \textit{laptop}) may appear in other labeled actions in the same video. Inspired by this observation, we propose to alleviate the missing label issue by exploiting the rich temporal correlations of object instances in the video. The core idea is that action labels in a different period may help to infer the presence of some objects in this current period. Specifically, a novel temporal dynamic graph LSTM (TD-Graph LSTM) framework is introduced to model the complex and dynamic temporal graph structure for object proposals in the whole video and thus enable the joint reasoning for all frames. The knowledge of all action labels in the video can thus be effectively transfered into all frames to enhance their frame-level categorizations.

To incorporate the temporal correlation of object proposals for global reasoning, we resort to the family of recurrent neural networks \cite{hochreiter1997long} due to their good sequential modeling capability. However, existing recurrent networks are largely limited in the constrained information propagation on fixed nodes following predefined routes such as tree-LSTM  \cite{tai2015improved}, graph-LSTM \cite{liang2016semantic} and structural-RNN \cite{JainZSS16, xiaodancvpr17}. In contrast, due to the unknown object localizations and temporal motion, it is difficult to find an optimal structure that connects object proposals for routed information propagation to achieve weakly-supervised video object detection. The proposed TD-Graph LSTM, posed as a general dynamic recurrent structure, overcomes these limitations by performing the dynamic information propagation based on an adaptive temporal graph that varies over both time periods in the video and model status in each updating step. 

Specifically, the dynamic temporal graph is constructed based on the visual correlation of object proposals across neighboring frames. The set of graph nodes denotes the entire collection of object proposals in all the frames, while graph edges are adaptively specified for consecutive frames in distinct learning steps. At each iteration, given the updated feature representation of object proposals, we only activate the edge connections with object proposals that have highest similarities with each current proposal. The adaptive graph topology can thus be constructed where different proposals are connected with different temporal correlated neighbors. TD-Graph LSTM alternatively performs the information propagation through each temporal graph topology and updates the graph topology at each iteration. In this way, our model enables the joint optimization of feature learning and temporal inference towards a robust slightly-supervised detection framework.

The contributions of this paper are summarized as 1) We explore a new slightly-supervised video object detection pipeline that leverages convenient action descriptions as the supervision; 2) A novel TD-Graph LSTM framework alleviates the missing label issue by enabling global reasoning over the whole video; 3) TD-Graph LSTM is posed as a general dynamic recurrent structure that performs temporal information propagation on an adaptively updated graph topology at each iteration; 4) We collect and release 5,000 frame annotations with object-level bounding boxes on daily-life videos, with the goal of evaluating our model and also helping advance the object detection community.

\begin{figure*}[!tp]
  	\begin{center}
  		\includegraphics[scale=0.42]{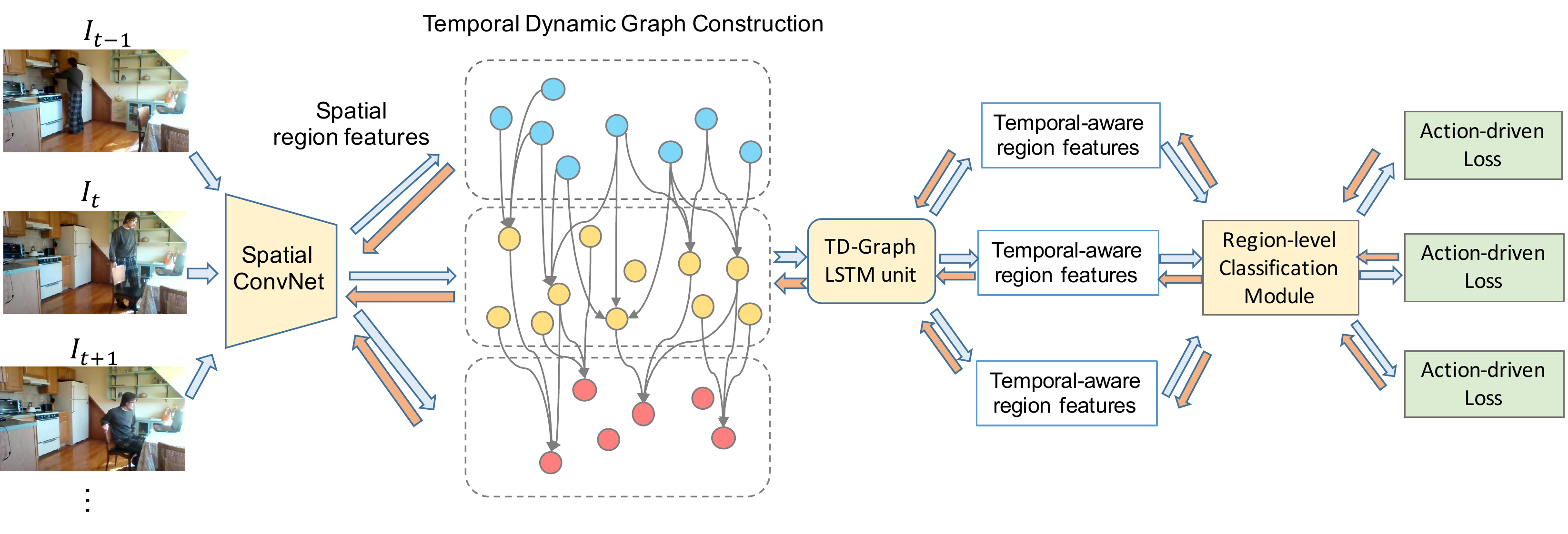}
  		\caption{{Our TD-Graph LSTM. Each frame is first passed into a spatial ConvNet to extract region-level features. A temporal graph structure is then constructed by dynamic edge connections between regions in two consecutive frames. TD-Graph LSTM then recurrently propagates information over the updated graph to generate temporal-aware feature representations for all regions. A region-level classification module is then adopted to produce category confidences of all regions in each frame, which are aggregated to obtain frame-level action predictions. The final action-driven loss for each frame is used to feedback signals into the whole model. After each gradient updating, the temporal graph is dynamically updated based on new visual features. For clarity, some edges in the graph are omitted.}}
  		\label{fig:framework}
  		\vspace{-6mm}
  	\end{center}
  \end{figure*}
  
\section{Related Works}
\textbf{Weakly-Supervised Object Detection.} Though recent state-of-the-art fully-supervised detection pipelines~\cite{he2015spatial,ren2015faster, girshick15fastrcnn, yolo16, ssd16} have achieved great progress, they heavily rely on large-scale bounding-box annotations. To alleviate this expensive annotation labor, weakly-supervised methods~\cite{deselaers2010localizing, siva2012defence, bilen2014weakly, song2014learning, wang2014weakly, cinbis2015multifold, Zequn17, zhang2017co} have recently attracted a lot of interest. These approaches use cheaper image-level object labels rather than bounding boxes. 
Beyond the image domain, another line of research~\cite{xiaolong15, xiaodan15,Krishna16,Alessandro12, Anestis13, Armand14, Suha15, zhang2017revealing} attempts to exploit the temporal information embedded in videos to facilitate the weakly-supervised object detection. Different from all the existing pipelines, we investigate a much cheaper action-driven object detection setting that aims to detect all object instances given only action descriptions. In addition, instead of employing multiple separate steps (e.g., detection and tracking) ~\cite{kai16,Suha15,xiaolong15,xiaodan15,Krishna16} to capture motion patterns, our TD-graph LSTM is an end-to-end framework that incorporates the intrinsic temporal coherence with a designed dynamic recurrent network structure into the action-driven slightly-supervised detection.

\textbf{Sequential Modeling.} Recurrent neural networks, especially Long Short-Term Memory (LSTM)~\cite{hochreiter1997long}, have been adopted to address many video processing tasks such as action recognition~\cite{NgHVVMT15}, action detection~\cite{YeungRMF16}, video prediction~\cite{SrivastavaMS15, ShiCWYWW15}, and video summarization \cite{zhang2016video}. 
However, limited by the fixed propagation route of existing LSTM structures~\cite{hochreiter1997long}, most of the previous works~\cite{NgHVVMT15, YeungRMF16, SrivastavaMS15} can only learn the temporal interdependency between the holistic frames rather than more fine-grained object-level motion patterns. Some recent approaches develop more complicated recurrent network structures. For instance, structural-RNN~\cite{JainZSS16} develops a scalable method for casting an arbitrary spatio-temporal graph as a rich RNN mixture.  A more recent Graph LSTM~\cite{liang2015semantic} defined over a pre-defined graph topology enables the inference for more complex structured data. However, both of them require a pre-fixed network structure for information propagation, which is impractical for weakly-supervised/slightly-supervised object detection without the knowledge of object localizations and precise object class labels. To handle the propagation over dynamically specified graph structures, we thus propose a new temporal dynamic network structure that supports the inference over the constantly changing graph topologies in different training steps. 

\section{The proposed TD-Graph LSTM}
\textbf{Overview.} We establish a fully-differentiable temporal dynamic graph LSTM (TD-Graph LSTM) framework for the action-driven video object detection task. For each video, the provided annotations are a set of action labels $\mathbf{Y} = \{y_1, \dots, y_N\}$, each of which describes the action $y_i = <a_i, c_i>$ appearing within a consecutive sequence of frames $\{I_{d^s_i}, \dots, I_{d_i^e}\}$, where $d^s_i$ and $d^e_i$ indicate the action starting and ending frame index. $a_i$ denotes the corresponding action noun while $c_i$ denotes the object noun. For example, the action \emph{tidying a shelf} is comprised of the action \emph{Tidying} and object \emph{a shelf}. To achieve weakly-supervised object detection, we only extract the object nouns $\{c_i\}$ of action labels in all videos and eliminate the prepositions (e.g., \emph{a}, \emph{the}) to produce an object category corpus (e.g., \emph{shelf}, \emph{door}, \emph{cup}) with $C$ classes. Each frame $I$ can be thus assigned with several participating object classes. For example, frames with two actions will be assigned with more than one participating object class, as shown in Figure~\ref{fig:problem}. The action-driven object detection is thus posed as a multi-class weakly-supervised video object detection problem. For simplicity, we eliminate the subscript $i$ of action labels in the following.

  \begin{figure*}[!tp]
  	\begin{center}
  		\includegraphics[scale = 0.45]{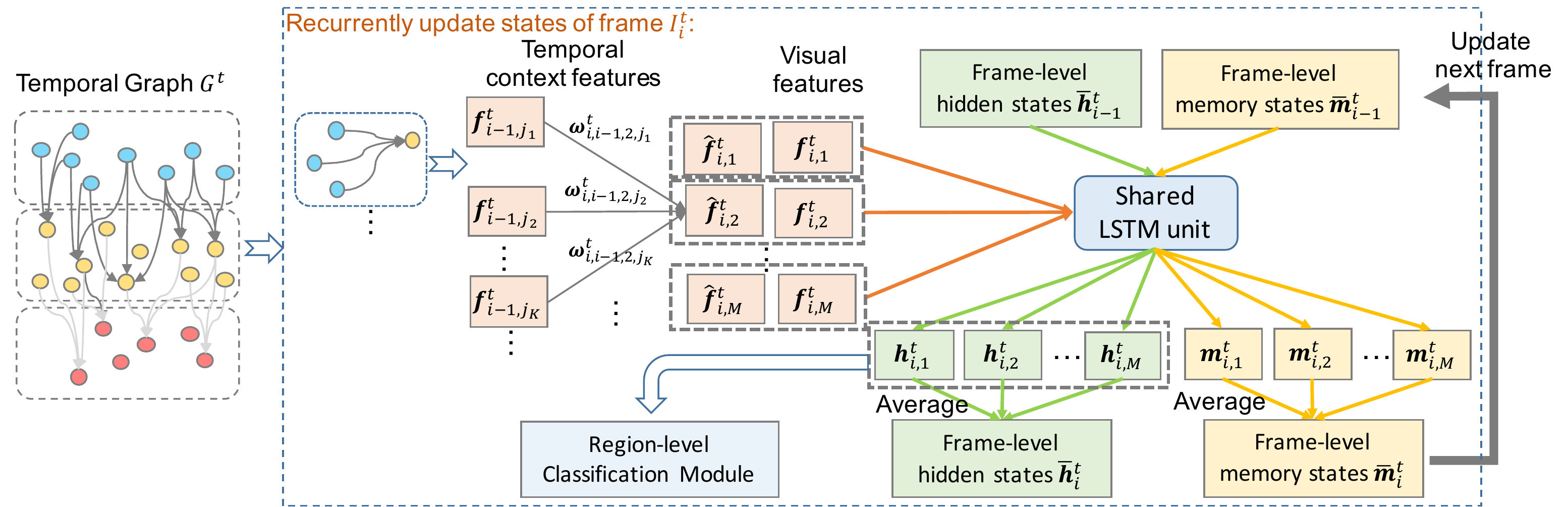}
  		\caption{{Illustration of the TD-Graph LSTM layer at $t$-th gradient updating. Given the constructed temporal graph $\mathcal{G}^t$, the TD-Graph LSTM recurrently updates the hidden states of each frame $I_i, i\in \{1,\dots, N\}$ as the enhanced temporal-aware visual feature, and then feeds these features into a region-level classification module to compute final category confidences of all regions. Specially, each LSTM unit takes the shared frame-level hidden states $\bar{\mathbf{h}}_{i-1}^t$ and memory states $\bar{\mathbf{m}}_{i-1}^t$, and input features for all regions as the inputs. Then the updated hidden states and memory states for all regions are produced, which are then averaged to generate the new frame-level hidden states $\bar{\mathbf{h}}_{i}^t$ and memory states $\bar{\mathbf{m}}_{i}^t$  for updating next frame $I_{i+1}$. The input features of each region consist of the visual features $\mathbf{f}_{i,j}^t$ and temporal context features $\hat{\mathbf{f}}_{i,j}^t$ that are aggregated by its connected regions with edge weights in the preceding frame.}}
  		\label{fig:TDGraphLSTM}
  		\vspace{-6mm}
  	\end{center}
  \end{figure*}
  
Figure~\ref{fig:framework} gives an overview of our TD-Graph LSTM. Each frame in the input video is first passed through a spatial ConvNet to obtain spatial visual features for region proposals. Based on visual features, similar regions in two consecutive frames are discovered and associated to indicate the same object across the temporal domain. A temporal graph structure is constructed by connecting all of the semantically similar regions in two consecutive frames, where graph nodes are represented by region proposals. The TD-Graph LSTM unit is then employed to recurrently propagate information over the whole temporal graph, where LSTM units take the spatial visual features as the input states. Benefiting from the graph topology, TD-Graph LSTM is capable of incorporating temporal motion patterns for participating objects in the action in a more efficient and meaningful way. TD-Graph LSTM outputs the enhanced temporal-aware features of all regions. Region-level classification is then employed to produce classification confidences. These region-level predictions can finally be aggregated to generate frame-level object class prediction, supervised by the object classes from action labels. The action-driven object categorization loss thus enables the holistic back-propagation into all regions in the video, where the prediction of each frame can mutually benefit from each other. 
\subsection{TD-Graph LSTM Optimization}
The proposed TD-Graph LSTM is comprised by three parametrized modules: spatial ConvNet $\Phi(\cdot)$ for visual feature extraction, TD-Graph LSTM unit $\Psi(\cdot)$ for recurrent temporal information propagation, and region-level classification module $\varphi(\cdot)$. These three modules are iteratively updated, targeted at the action-driven object detection.

At each model updating step $t$, a temporal graph structure $\mathcal{G}^t = <\mathbf{V}, \mathcal{E}^t>$ for each video is constructed based on the updated spatial visual features $\mathbf{f}^t$ of all regions $\mathbf{r}$ in the videos, defined as $ \mathcal{G}^t = \beta(\Phi^t(\mathbf{r}))$. $\beta(\cdot)$ is a function to calculate the dynamic edge connections $\mathcal{E}^t$ conditioning on the updated visual features $\mathbf{f}^t = \Phi^t(\mathbf{r})$. The TD-Graph LSTM unit $\Psi^t$ recurrently functions on the visual features $\mathbf{f}^t$ of all frames and propagates temporal information over the graph $ \mathcal{G}^t$ to obtain the enhanced temporal-aware features $\hat{\mathbf{f}}^t = \Psi^t(\mathbf{f}^t|\mathcal{G}^t)$ of all regions in the video. Based on the enhanced $\hat{\mathbf{f}}^t$, the region-level classification module $\varphi$ produces classification confidences  $\mathbf{rc}^t$ for all regions, as $\mathbf{rc}^t = \varphi(\hat{\mathbf{f}}^t)$. These region-level category confidences $\mathbf{rc}^t$ can be aggregated to produce frame-level category confidences $\mathbf{pc}^t = \gamma(\mathbf{rc}^t)$ of all frames by summing the category confidences of all regions of each frame.

During training, we define the action-driven loss for each frame as a hinge loss function and train a multi-label image classification objective for all frames in the videos:
\begin{equation}
	\begin{split}
&\mathcal{L}(\Phi, \Psi, \varphi)  = \dfrac{1}{CN}\sum_{c=1}^{C} \sum_{i=1}^{N}\max(0, 1-\mathbf{y}_{c,i}\mathbf{pc}_{c,i})\\
&= \dfrac{1}{CN}\sum_{c=1}^{C} \sum_{i=1}^{N} \max(0, 1-\mathbf{y}_{c,i}\gamma(\varphi(\Psi(\mathbf{f}_i|\mathcal{G})))),
	\end{split} 
	\label{eq:obj}
\end{equation}
where $ C $ is the number of classes and $ \mathbf{y}_{c,i}, i \in\{1,\dots, N\}$ represents action-driven object labels for each frame. For each frame $I_i$, $y_{c,i} = 1$ only if the action-driven object label $c$ is assigned to the frame $I_i$, otherwise as -1. The objective function defined in Eq.~\ref*{eq:obj} can be optimized by the Stochastic Gradient Descent (SGD) back-propagation. At each $t$-th gradient updating, the temporal graph structure $\mathcal{G}^t$ is accordingly updated by $\beta(\Phi^t(\mathbf{r}))$ for each video. Thus, the TD-Graph LSTM unit optimizes over a dynamically updated graph structure $\mathcal{G}^t$. In the following sections, we introduce the above-defined parametrized modules.
\subsection{Spatial ConvNet}

Given each frame $I_i$, we first extract category-agnostic region proposals and then extract their visual features by passing them into a spatial ConvNet $\Phi(\cdot)$ following~\cite{girshick15fastrcnn}. To provide a fair comparison on action-driven object detection, we adopt the EdgeBoxes~\cite{Uijlings13} proposal generation method which does not require any object annotations for pretraining. We select the top $M = 500$ proposals $ \mathbf{r}_{i} = \{r_{i,1}, r_{i,2}, ..., r_{i,M} \}$ for the frame $I_i$ with the highest objectness scores, considering the computation efficiency. At the $t$-th updating step, visual features $ \mathbf{f}^t_{i} = \{f^t_{i,1}, f^t_{i,2}, ..., f^t_{i,M}\} \in \mathbb{R}^{M\times D}$ of all regions $\mathbf{r}_i$ are extracted using the updated spatial ConvNet model, i.e., $\mathbf{f}^t_i = \Phi^t(\mathbf{r}_i)$. The spatial ConvNet $\Phi(\cdot)$ consists of several convolutional layers from the base net and one ROI-pooling layer \cite{girshick15fastrcnn}, and two fully-connected layers. 
  
\subsection{TD-Graph LSTM Unit}

\textbf{Dynamic Graph Updating.} Given the updated visual features $\mathbf{f}^t_i$ of each frame $I_i$, the temporal graph structure $\mathcal{G}^t = <\mathbf{V}, \mathcal{E}^t>$ can be accordingly constructed by learning the dynamic edge connections $\mathcal{E}^t$. The graph node $\mathbf{V} = \{v_{i,j}\}, j = \{1,\dots,M\}$ is represented by visual features $\{\mathbf{f}^t_{i,j}\}$ of all regions in all frames; that is, $M\times N$ nodes for $M$ region proposals of $N$ frames. 
Each node $v_{i,j}$ is connected with nodes in the preceding frame $I_{i-1}$ and the nodes in subsequent frame $I_{i+1}$. To incorporate the motion dependency in consecutive frames, the edge connections $\mathcal{E}^t_{i,i-1}$ between nodes in $I_i$ and $I_{i-1}$ are mined by considering their appearance similarities in visual features. Specifically, the edge weight between each pair of nodes $(v_{i,j}, v_{i-1,j'})$ is first calculated as $\frac{1}{2}\exp(-||\mathbf{f}^t_{i,j} - \mathbf{f}^t_{i-1,j'}||_2)$. To make the model inference efficient and alleviate the missing issue, each node $v_{i,j}$ is only connected to $K$ nodes $v_{i-1,j'}$ with the top-$ K $ highest edge weights in preceding frame $I_{i-1}$, and these activated edge weights are normalized to be summed as 1. We denote the normalized edge weight as $\omega^t_{i,i-1,j,j'}$. Thus, the updated temporal graph structure $\mathcal{G}^t$ can be regarded as an undirected $K$-neighbor graph where each node $v_{i,j}$ is connected with at most $K$ nodes in previous frames. 

\textbf{TD-Graph LSTM.} TD-Graph LSTM layer propagates temporal context over graph and recurrently updates the hidden states $\{\mathbf{h}_{i,j}^t\}$ of all regions in each frame $I_i$ to construct enhanced temporal-aware feature representations.
These features are fed into the region-level classification module to compute the category-level confidences of each region. TD-Graph LSTM updates hidden state of frame $i$ by incorporating information from frame-level hidden state $\bar{\mathbf{h}}_{i-1}^t$ and memory state $\bar{\mathbf{m}}_{i-1}^t$.
The usage of the shared frame-level hidden state and memory state enables the provision of a compact memorization of temporal patterns in the previous frame and is more suitable for massive and possibly missing graph nodes (e.g., 500 in our setting) in a large temporal graph. After performing $N$ updating steps for all frames, our model effectively embeds the rich temporal dependency to obtain the enhanced temporal-aware feature representations of all regions in all frames. For updating the features of each node $v_{i,j}$ in the frame $I_i$, the TD-Graph LSTM unit takes as the input its own visual features $\mathbf{f}_{i,j}^t$, temporal context features $\hat{\mathbf{f}}_{i,j}^t$, frame-level hidden states $\bar{\mathbf{h}}_{i-1}^t$ and memory states $\bar{\mathbf{m}}_{i-1}^t$, and outputs the new hidden states $\mathbf{h}_{i,j}^t$. Given the dynamic edge connections $e_{i,j} = \{<v_{i,j}, v_{i-1,j'}>\}, j' \in \mathcal{N}_{\mathcal{G}}(v_{i,j})$, the temporal context features $\hat{\mathbf{f}}_{i,j}^t$ can be calculated by performing a weighted summation of features of connected regions:
\begin{equation}
\hat{\mathbf{f}}_{i,j}^t = \sum_{j'\in\mathcal{N}_{\mathcal{G}}(v_{i,j})}\omega^t_{i,i-1,j,j'}\mathbf{f}^t_{i-1,j'}.\vspace{-3mm}
\end{equation}
And the shared frame-level hidden states $\bar{\mathbf{h}}_{i-1}^t$ and memory states $\bar{\mathbf{m}}_{i-1}^t$ can be computed as
\begin{equation}
\bar{\mathbf{h}}_{i-1}^t = \frac{1}{M}\sum_{j=1}^{M}\mathbf{h}_{i-1,j}^t, \quad \bar{\mathbf{m}}_{i-1}^t = \frac{1}{M}\sum_{j=1}^{M}\mathbf{m}_{i-1,j}^t.
\end{equation}
The TD-Graph LSTM unit consists of four gates for each node $v_{i,j}$:  the input gate $\mathbf{gu}^t_{i,j}$, the forget gate $\mathbf{gf}^t_{i,j}$, the memory gate $\mathbf{gc}^t_{i,j}$, and the output gate $\mathbf{go}^t_{i,j}$. The $W^u_t, W^f_t, W^c_t, W^o_t$ are the recurrent gate weight matrices specified for input visual features and  $W^{ut}_t,W^{ft}_t, W^{ct}_t, W^{ot}_t$ are those for temporal context features. $U^{u}_t, U^{f}_t, U^{c}_t, U^{o}_t$ are the weight parameters specified for frame-level hidden states. The new hidden states and memory states in the graph $\mathcal{G}^t$ can be calculated as follows:
\begin{equation}
	\begin{split}
	\mathbf{gu}^t_{i,j} = &\delta(W^u_t\mathbf{f}_{i,j}^{t} + W^{ut}_t\hat{\mathbf{f}}_{i,j}^{t} + U^{u}_t\bar{\mathbf{h}}_{i-1}^{t} + b^u_t),\\
	\mathbf{gf}^t_{i,j} = &\delta(W^f_t\mathbf{f}_{i,j}^{t} + W^{ft}_t\hat{\mathbf{f}}_{i,j}^{t} + U^f_t\bar{\mathbf{h}}_{i-1}^{t} + b^f_t),\\
	\mathbf{go}^t_{i,j} = &\delta(W^o_t\mathbf{f}_{i,j}^{t} + W^{ot}_t\hat{\mathbf{f}}_{i,j}^{t} + U^{o}_t\bar{\mathbf{h}}_{i-1}^{t} + b^o_t),\\
	\mathbf{gc}^t_{i,j} = &\tanh(W^c_t\mathbf{f}_{i,j}^{t}  + W^{ct}_t\hat{\mathbf{f}}_{i,j}^{t} + U^c_t\bar{\mathbf{h}}_{i-1}^{t} + b^c_t),\\
	\mathbf{m}^t_{i,j} = &\mathbf{gf}^t_{i,j} \odot \bar{\mathbf{m}}_{i-1}^{t} + \mathbf{gu}^t_{i,j}\odot \mathbf{gc}^t_{i,j},\\
	\mathbf{h}^t_{i,j} = &\mathbf{go}^t_{i,j}\odot\tanh(\mathbf{m}^t_{i,j}).
	\end{split}
	\label{eq:lstm}
	\end{equation}

\noindent{Here} $\delta$ is a logistic sigmoid function, and $\odot$ indicates a point-wise product. Given the updated hidden states $\{\mathbf{h}^t_{i,j}\}$ and memory states $\{\mathbf{m}^t_{i,j}\}$ of all regions in frame $I_i$, we can obtain new frame-level hidden states $\bar{\mathbf{h}}_{i}^{t}$ and memory states $\bar{\mathbf{m}}_{i}^{t}$ for updating the states of regions in frame $I_{i+1}$. The TD-LSTM unit recurrently updates the states of all regions in each frame, and thus the past temporal information in preceding frames can be utilized for updating each frame. The TD-Graph LSTM layer is illustrated in Figure~\ref{fig:TDGraphLSTM}. 

\subsection{Region-level Classification Module}
Given the updated hidden states $\mathbf{h}_{i,j}^t$ for each node $v_{i,j}$, we use a region-level classification module to obtain the category confidences of all regions, that is, $\mathbf{rc}_{i}^t = \varphi({\mathbf{h}}_{i}^t)$ of all $M$ regions. Following the two-stream architecture of WSDDN~\cite{bilen2016weakly}, the region-level classification module contains a \textit{detection stream} and a \textit{classification stream}, and produces final classification scores by performing element-wise multiplication between them. The \textit{classification stream} takes the region-level feature vectors $\mathbf{h}_{i}^t$ of all regions as the input and feeds it to a linear layer that outputs a set of class scores $ \mathbf{S}_{i}^t\in \mathbb{R}^{M\times C}$ for $C$ classes of all $M$ regions. Here, we use the reproduced WSDDN in~\cite{kantorov2016contextlocnet} that does not employ an additional softmax in the \textit{classification stream}. These differences have a minor effect on the detection accuracy as has been discussed in~\cite{kantorov2016contextlocnet}. The \textit{detection stream} also takes $ \mathbf{h}_{i}^t$ as the input and feeds it to another linear layer that outputs a set of class scores, giving a matrix of scores $ \mathbf{L}_i^t \in \mathbb{R}^{M\times C}$. $ \mathbf{L}_i^t $ is then fed to another softmax layer to normalize the scores over the regions in the frame. The final scores of all regions $\mathbf{rc}_{i}^t$ are obtained by taking the element-wise multiplication of the two scoring matrices $ \mathbf{S}_{i}^t$ and $ \mathbf{L}_i^t$. We sum all the region-level class scores $\mathbf{rc}_{i}^t$ to obtain the frame-level class prediction scores $\mathbf{pc}_{i}^t$.


\section{Experiments}
\subsection{Dataset and Evaluation Measures}
\begin{figure*}[!tp]
  	\begin{center}
  		\includegraphics[scale = 0.55]{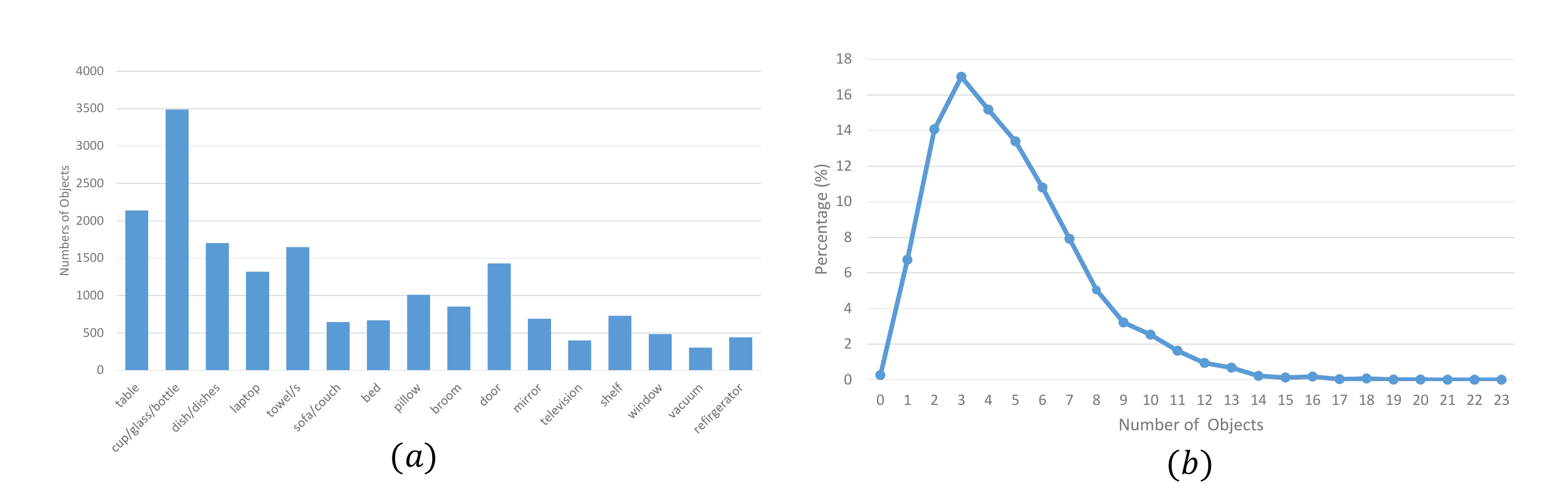}
  		\caption{{(a) The distribution of object classes appearing in the action labels of the training set. (b) The distribution of the ground truth bounding box numbers in each image of the test set. }}
  		\label{fig:datasetdistribution}
  		\vspace{-4mm}
  	\end{center}
  \end{figure*}

\textbf{Dataset Analysis.} We evaluate the action-drive weakly-supervised object detection performance on the \textit{Charades} dataset~\cite{SigurdssonVWFLG16}. The \textit{Charades} video dataset is composed of daily indoor activities collected through Amazon Mechanical Turk. There are 157 action classes and on average 6.8 actions in each video, which occur in various orders and contexts. In order to detect objects in videos by using action labels, we only consider the action labels that are related to objects for training. Therefore, there are 66 action labels that are related to 17 object classes in our experiments. We show distribution of object classes (in a random subset of videos) in Figure~\ref{fig:datasetdistribution} (a). The training set contains 7,542 videos. Videos are down-sampled to 1 fps and we only sample the frames assigned with action labels in each video. During training, only frame-level action labels are provided for each video.

\begin{figure}[!tp]
  	\begin{center}
  		\includegraphics[scale = 0.32]{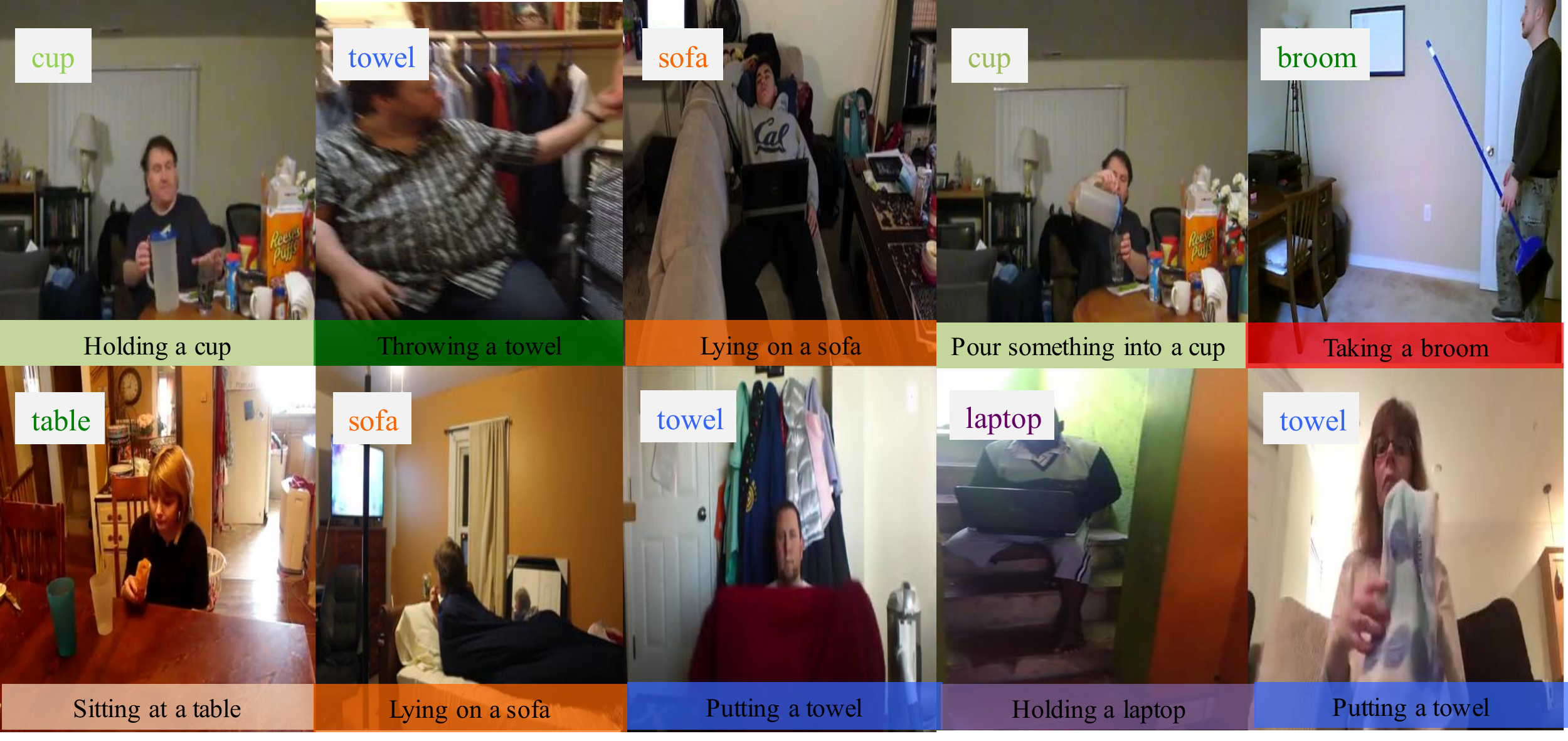}
  		\caption{{Several samples of key frames from videos in \textit{Charades}. The action labels are given at the bottom of the image and the related objects are listed at the top of the image.}}
  		\label{fig:datasetsample}
  		\vspace{-6mm}
  	\end{center}
  \end{figure}

In order to evaluate the video object detection performance over 17 daily object classes, we collect the bounding box annotations for 5,000 test frames from 200 videos in the Charades test set. The bounding box number distribution in each frame is shown in Figure~\ref{fig:datasetdistribution} (b), ranging from 1 to 23 boxes appearing in the frame. 
More than 60\% frames have more than 4 bounding boxes and most video frames exhibit severe motion blurs and low resolution. This poses more challenges for the object detection model compared to an image-based object detection dataset, such as the most popular PASCAL VOC~\cite{everingham2010pascal} that is widely used in existing weakly-based object detection methods. Figure~\ref{fig:datasetsample} further shows example frames with action labels on the \textit{Charades} dataset. It can be seen that each action label only provides one piece of object class information for the frame that may contain several object classes, which can be regarded as the missing label issue for training a model under this action-driven setting. Moreover, the video frames often appear with a very cluttered background, blurry objects and diverse viewpoints, which are more challenging and realistic compared to existing image datasets (e.g., MS COCO{\cite{Lin2014}} and ImageNet{\cite{imagenet}}) and video datasets (e.g., UCF101{\cite{UCF101}}).

\textbf{Evaluation Measures.} We evaluate the performance of both object detection and image classification tasks on \textit{Charades}. For detection, we report the average precision (AP) at 50\% intersection-over-union (IOU) of the detected boxes with the ground truth boxes. For classification, we also report the AP on frame-level object classification. 

\begin{table*}[!tp]\setlength{\tabcolsep}{1.5pt}
    \small
	\centering\scriptsize
	\caption{Per-class performance comparison of our proposed models with two state-of-the-art weakly-supervised learning methods when evaluating on the \textit{Charades} dataset\cite{SigurdssonVWFLG16}, test \textbf{classification} average precision (\%).}\label{tab:classmAP}
	\begin{tabular}{cccccccccccccccccccccccccccccc}
		\toprule
		Method &  bed  &  broom  & chair & cup & dish & door & laptop & mirror & pillow & refri & shelf & sofa & table & tv	& towel &	vacuum & window & mAP\\
		\midrule
		WSDDN~\cite{bilen2016weakly}& 39.8 & 5.85 & 36.1 & 21 & 16.3 & 11.6 & 30.5 & 4.7 & 2.8 & 6.5 & 8.1 & 14.8 & 37.8 & 5 & 12.5 & 8.2 & 4.8 & 15.67\\
		ContextLocNet~\cite{kantorov2016contextlocnet} & 43.37 & 5.65 & 38.95 & 16.62 & 12.46 & 8.67 & 27.75 & 4.5 & 3.51 & \textbf{11.12} & \textbf{9.79} & 15.67 & 37.44 & \textbf{14.39 }& 9.72 & 16.36 & 3.97 & 16.47 \\
		\midrule
		{TD-Graph LSTM w/o LSTM} &  {32.54} & {5.875} & {31.69} & {\textbf{27.9}} & {15.79} & {14.19} & {18.81} & {\textbf{6.15}} & {8.35} & {4.5} & {9.3} & {24.33} & {33} & {8.26} & {14.7} & {7.68} & {\textbf{6.72}} & {15.89}\\
		{TD-Graph LSTM w/o graph} & {25.04} & {6.51} & {43.79} & {21.54} & {15.6} & {\textbf{15.86}} & {19.57} & {5.61} & {9.32} & {6.2} & {9.02} & {\textbf{25.95}} & {39.2} & {8.85} & {\textbf{15.27}} & {\textbf{18.18}} & {5.63} & {17.13}\\
		{\textbf{TD-Graph LSTM}} & {\textbf{47.62}} & {\textbf{12.26}} & {\textbf{45.07}} & {23.55} & {\textbf{16.7}} & {15.6} & {\textbf{30.9}} & {5.05} & {\textbf{17.64}} & {7.43} & {9.53} & {19.52} & {\textbf{43.29}} & {4.23} & {12.47} & {15.03} & {5.91} & {\textbf{19.52}}\\	
		\bottomrule
		\vspace{-4mm}
	\end{tabular}%
\end{table*}%

\begin{table*}[!tp]\setlength{\tabcolsep}{1.5pt}
    \small
	\centering\scriptsize
	\caption{Per-class performance comparison of our proposed models with two state-of-the-art weakly-supervised learning methods when evaluating on the \textit{Charades} dataset\cite{SigurdssonVWFLG16}, test \textbf{detection} average precision (\%).}\label{tab:detectmAP}
	\begin{tabular}{cccccccccccccccccccccccccccccc}
		\toprule
		Method &  bed  &  broom  & chair & cup & dish & door & laptop & mirror & pillow & refri & shelf & sofa & table & tv	& towel &	vacuum & window & mAP\\
		\midrule
		WSDDN~\cite{bilen2016weakly}& {2.38} & {0.04} & {1.17} & {0.03} & {\textbf{0.13}} & {0.31} & {2.81} & {0.28} & {0.02} & {0.12} & {0.03} & {0.41} & {1.74} & {1.18} & {0.07} & {0.08} & {0.22} & {0.65}  \\
		ContextLocNet~\cite{kantorov2016contextlocnet} & {7.4} & {0.03} & {0.55} & {0.02} & {0.01} & {0.17} & {1.11} & {0.66} & {0} & {0.07} & {1.75} & {4.12} & {0.63} & {0.99} & {0.03} & {\textbf{0.75}} & {0.78} & {1.12} \\
		\midrule
        {TD-Graph LSTM w/o LSTM} & {7.41} & {\textbf{0.05}} & {3} & {0.05} & {0.02} & {0.56} & {0.11} & {0.65} & {0.04} & {0.16} & {0.25} & {1.67} & {2.46} & {1.24} & {\textbf{0.11}} & {0.46} & {\textbf{1.46}} & {1.16}\\
		{TD-Graph LSTM w/o graph} & {\textbf{9.69}} & {0.02} & {2.85} & {0.34} & {0.05} & {0.87} & {1.95} & {\textbf{0.69}} & {0.05} & {\textbf{0.44}} & {2.11} & {3.34} & {1.91} & {1.05} & {0.05} & {0.29} & {0.69} & {1.55}\\
		{\textbf{TD-Graph LSTM}} & {9.19} & {0.04} & {\textbf{4.18}} & {\textbf{0.49}} & {0.11} & {\textbf{1.17}} & {\textbf{2.91}} & {0.3} & {\textbf{0.08}} & {0.29} & {\textbf{3.21}} & {\textbf{5.86}} & {\textbf{3.35}} & {\textbf{1.27}} & {0.09} & {0.6} & {0.47} & {\textbf{1.98}}\\
		\bottomrule
		\vspace{-6mm}
	\end{tabular}%
\end{table*}%

\subsection{Implementation Details}
Our TD-Graph LSTM adopts the VGG-CNN-F model~\cite{ChatfieldSVZ14} pre-trained on ImageNet ILSVRC 2012 challenge data~\cite{imagenet} as the base model, and replaces the last pooling layer $ pool5 $ with an SPP layer~\cite{he2015spatial} to be compatible with the first fully connected layer. 
We use the EdgeBoxes algorithm~\cite{ZitnickDollarECCV14edgeBoxes} to generate the top 500 regions that have width and height larger than 20 pixels as candidate regions for each frame. To balance the performance and time cost, we set the number of edges linked to each node $ K $ to 100. For training, we use stochastic gradient descent with momentum 0.9 and weight decay $ 5\times 10^{-4} $. All weight matrices used in the TD-Graph LSTM units are randomly initialized from a uniform distribution of $ [-0.1, 0.1] $. TD-Graph LSTM predicts the hidden and memory states with the same dimension as the previous region-level CNN features. Each mini-batch contains at most 6 consecutive sampled frames in a video. The network is trained on the \textit{Charades} training set by using fine-tuning on all layers, including those of the pre-trained base CNN model. The experiments are run for 30 epochs for the model convergence. The learning rates are set to $ 10^{-5} $ for the first ten epochs, then decreased to $ 10^{-6} $. All our models are implemented on the public Torch~\cite{Collobert_NIPSWORKSHOP_2011} platform, and all experiments are conducted on a single NVIDIA GeForce GTX TITAN X GPU with 12 GB memory. The runtime is 2.5 fps and 3.9 fps for training and testing respectively.


\begin{figure}[!tp]
  	\begin{center}
  		\includegraphics[scale = 0.5]{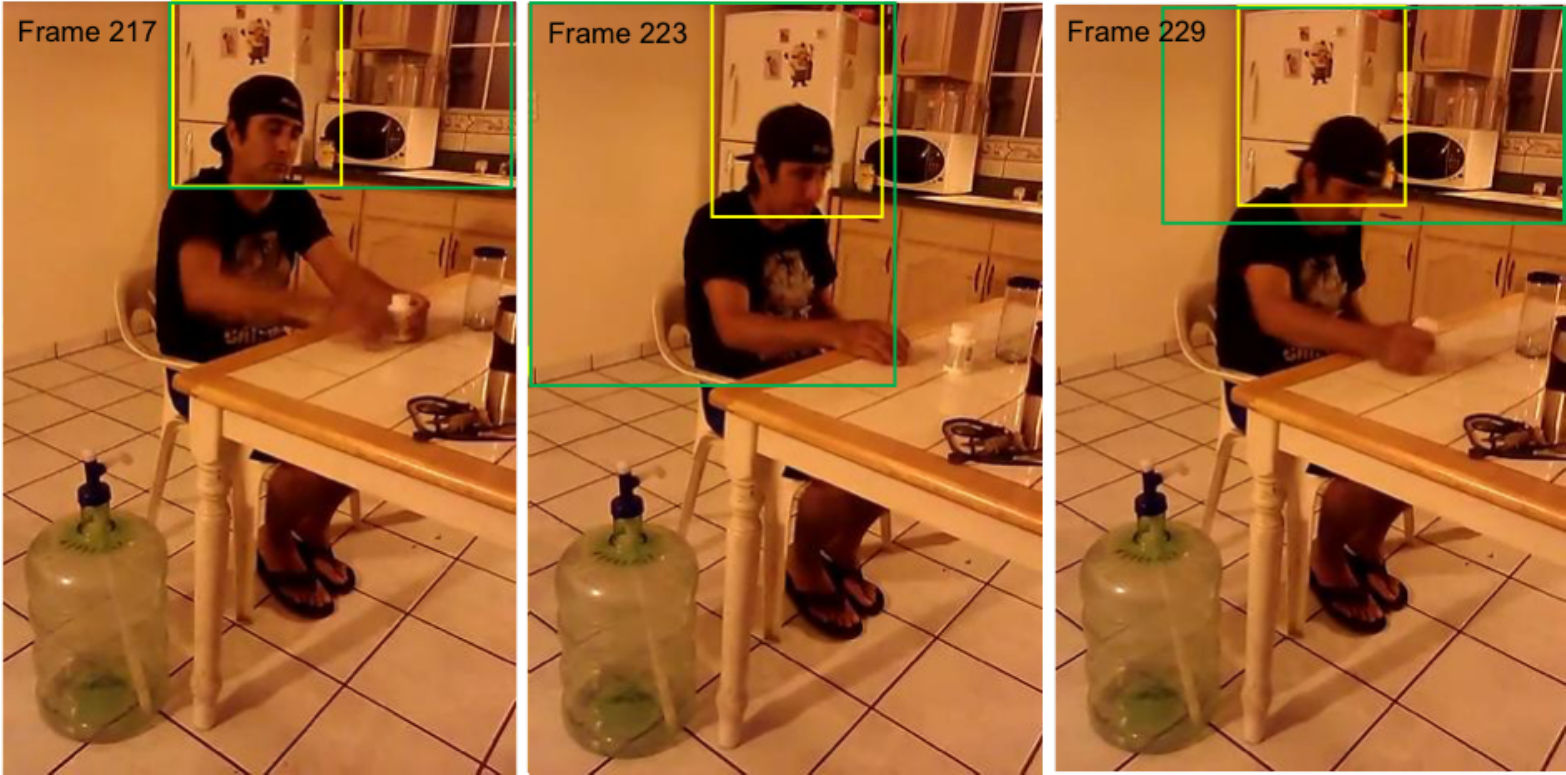}
  		\caption{{Our TD-Graph LSTM addresses well the missing label issue. It can successfully detect the refrigerator that is not referred to by any action labels (A green box shows the detection result and yellow box the ground truth.)}}
  		\label{fig:missinglabel}
  		\vspace{-8mm}
  	\end{center}
  \end{figure}
  
\subsection{Results and Comparisons}

\begin{figure*}[!tp]
  	\begin{center}
  		\includegraphics[scale = 0.43]{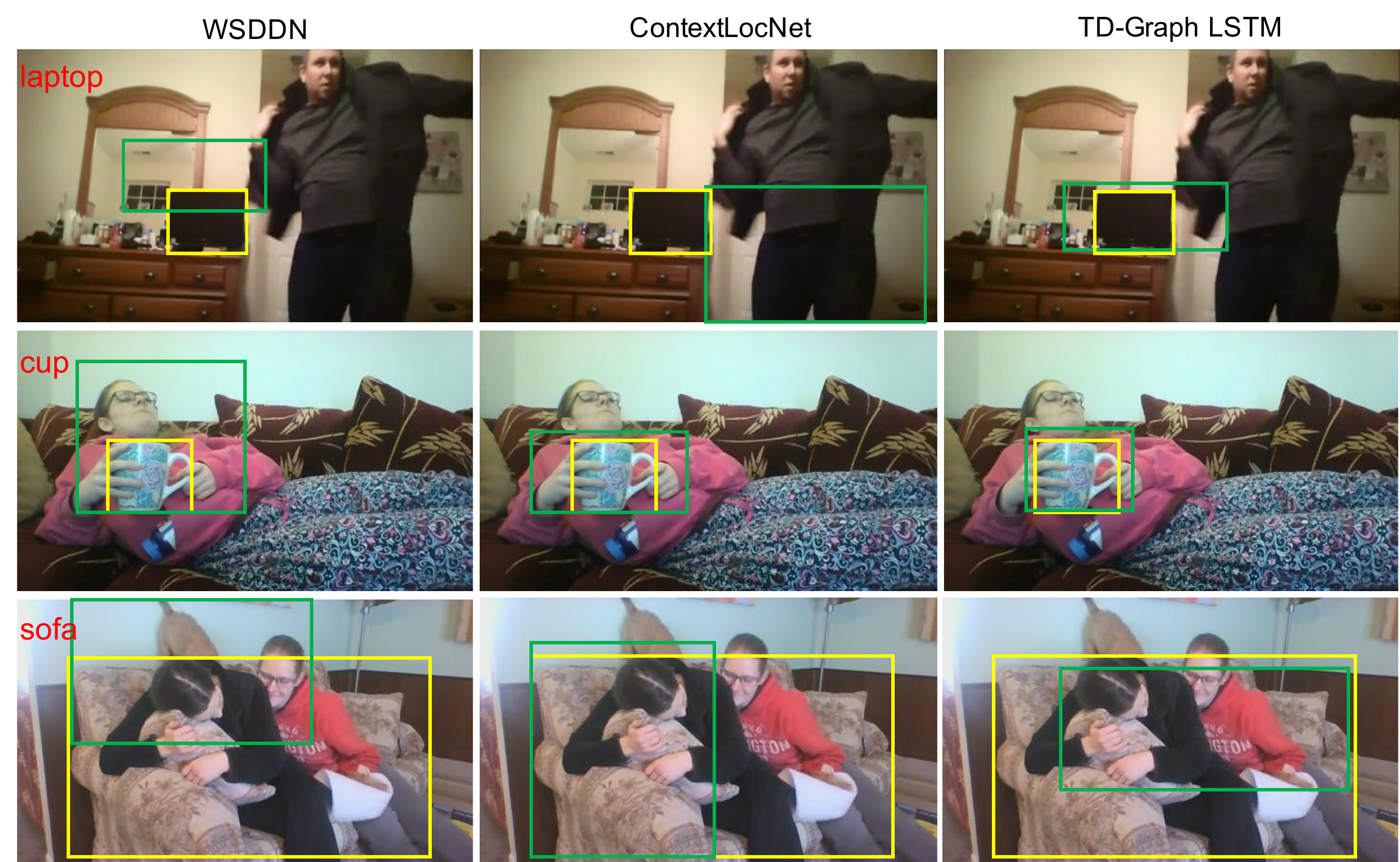}
  		\caption{{Qualitative comparisons with two state-of-the-arts on video object detection. The green boxes indicate detection results and yellow ones are the ground truth.}}
  		\label{fig:comparison}
  		\vspace{-6mm}
  	\end{center}
  \end{figure*}
  
We compare the proposed TD-Graph LSTM model with two state-of-the-art weakly-supervised learning methods on the \textit{Charades} dataset, WSDDN~\cite{bilen2016weakly} and ContextLocNet~\cite{kantorov2016contextlocnet}. As both of the two methods were proposed for image-based weakly-supervised image object detection, here we run the source code of ContextLocNet~\cite{kantorov2016contextlocnet} and their reproduced WSDDN\footnote{\url{https://github.com/vadimkantorov/contextlocnet}} on the \textit{Charades} dataset to make a fair comparison with our method. Their models are trained by treating the action-related object labels in each frame as the supervision information and are evaluated on each video frame. The difference between our model and WSDDN~\cite{bilen2016weakly} is our usage of TD-Graph LSTM layers to leverage rich temporal correlations in the whole video. Similar to WSDDN, ContextLocNet is also a two stream model with an enhanced localization module using various surrounding context. Specifically, we use the contrastive-S setup of ContextLocNet. All of these models use the same base model and region proposal method, i.e., VGG-CNN-F model~\cite{ChatfieldSVZ14} and EdgeBoxes~\cite{ZitnickDollarECCV14edgeBoxes}. 

We report the comparisons with two state-of-the-art on classification mAP and detection mAP in Table~\ref{tab:classmAP} and Table~\ref{tab:detectmAP}, respectively. It can be observed that our TD-Graph LSTM model substantially outperforms two baselines on both classification mAP and detection mAP, particularly, 3.05\% higher than ContextLocNet~\cite{kantorov2016contextlocnet} and 3.85\% than WSDDN~\cite{bilen2016weakly} in terms of classification mAP. Especially, our TD-Graph LSTM surpasses two baselines in small objects, e.g., over 14.13\% for pillow class and 6.93\% for cup class. Although our model and two baselines all obtain low detection mAP under this challenging setting, our TD-Graph LSTM still surpasses two baselines on detecting crowded and small objects in the video. The superiority of our TD-Graph LSTM clearly demonstrates its effectiveness in challenging action-driven weakly-supervised object detection where the missing label issue is quite severe and a considerable number of bounding boxes appear in each frame with very low quality. We further show the qualitative comparison with two state-of-the-arts in Figure~\ref{fig:comparison}. Our model is able to produce more precise object detection for even very small objects (e.g., the cup in the middle row) and objects with heavy occlusion (e.g., the sofa in the bottom row). Our TD-Graph LSTM takes the advantage of exploiting complex temporal correlations between region proposals by propagating knowledge into a whole dynamic temporal graph, which effectively alleviates the critical missing label issue, as shown in Figure~\ref{fig:missinglabel}.


\begin{table}[!tp]\setlength{\tabcolsep}{5pt}
	\centering\scriptsize
	\caption{Performance comparison of using different graph topologies when evaluating on the \textit{Charade}s dataset, test detection mAP (\%) and classification mAP (\%).}\label{tab:graph}
	\begin{tabular}{cccccccccccccccccccccccccccccc}
		\toprule
		Method &  det mAP  &  cls mAP \\
		\midrule
		{Ours w/o Graph} & {1.55} & {17.13} \\
		{Ours w/ Mean Graph} & {1.41} & {16.92} \\
		{Ours w/ Static Graph} & {1.89} & {17.97} \\
		{\textbf{Ours}} & {\textbf{1.98}} & {\textbf{19.52}}  \\
		\bottomrule
		\vspace{-8mm}
	\end{tabular}%
\end{table}%

\subsection{Ablation Study}
The results of model variants are reported in Table~\ref{tab:classmAP}, Table~\ref{tab:detectmAP} and Table~\ref{tab:graph}.

\textbf{The effectiveness of incorporating graph.} The main difference between our TD-Graph with a conventional LSTM structure for sequential modeling is in propagating information over a dynamic graph structure. To verify its effectiveness, we thus compare our full model with the variant ``TD-Graph LSTM w/o graph" that eliminates the edge connections between regions in consecutive frames, and updates the frame-level hidden and memory states with the original region-level features. Our TD-Graph LSTM consistently obtains better results over ``TD-Graph LSTM w/o graph", which speaks to the advantage of incorporating a graph for the challenging action-driven object detection. 

\textbf{The effectiveness of temporal LSTM.}
We further verify that recurrent sequential modeling by the LSTM units over the temporal graph is beneficial for exploiting complex object motion patterns in daily videos. ``TD-Graph LSTM w/o LSTM" indicates removing the LSTM units and directly aggregating the temporal context features to enhance features of each region. The performance gap between our full model and ``TD-Graph LSTM w/o LSTM" verifies the benefits of adopting LSTM. 

\textbf{Dynamic graph vs Static graph vs Mean graph.}
Besides the proposed dynamic graph, another commonly used alternative is the fully-connected graph where each region is densely connected with all regions in the preceding frame; that is, ``Ours w/ Static Graph" and ``Ours w/ Mean Graph". ``Ours w/ Static Graph" uses the adaptive edge weights similar to TD-Graph LSTM while ``Ours w/ Mean Graph" uses the same weights for all edge connections. It can be seen that applying a dynamic graph structure can help significantly boost both detection and classification performance over other fully-connected graphs. The reason is that meaningful temporal correlations between regions can be discovered by the dynamic graph and leveraged to transfer motion context into the whole video. 

\section{Conclusion}
In this paper, we propose a novel temporal dynamic graph LSTM architecture to address action-driven weakly-supervised object detection. It recurrently propagates the temporal context on a constructed dynamic graph structure for each frame. The global action knowledge in the whole video can be effectively leveraged for object detection in each frame, which helps alleviate the missing label problem. Extensive experiments on a large-scale daily-life action dataset \textit{Charades} demonstrate the superiority of our model over the state-of-the-arts. 

\noindent {\footnotesize{\bf Acknowledgements:} This work was supported by ONR MURI N000141612007 and Sloan Fellowship to AG. XL was supported by the Department of Defense under Contract No. FA8702-15-D-0002 with CMU for the operation of the Software Engineering Institute, a federally funded research and development center.
}

{\small
\bibliographystyle{ieee}
\bibliography{egbib}
}

\onecolumn
\centering{\section*{Supplementary Materials}}
\begin{flushleft}
We visualize more detection results of the TD-Graph LSTM on the Charades dataset.
\end{flushleft}

  \begin{figure*}[!h]
  	\begin{center}
  		\includegraphics[width=0.9\linewidth]{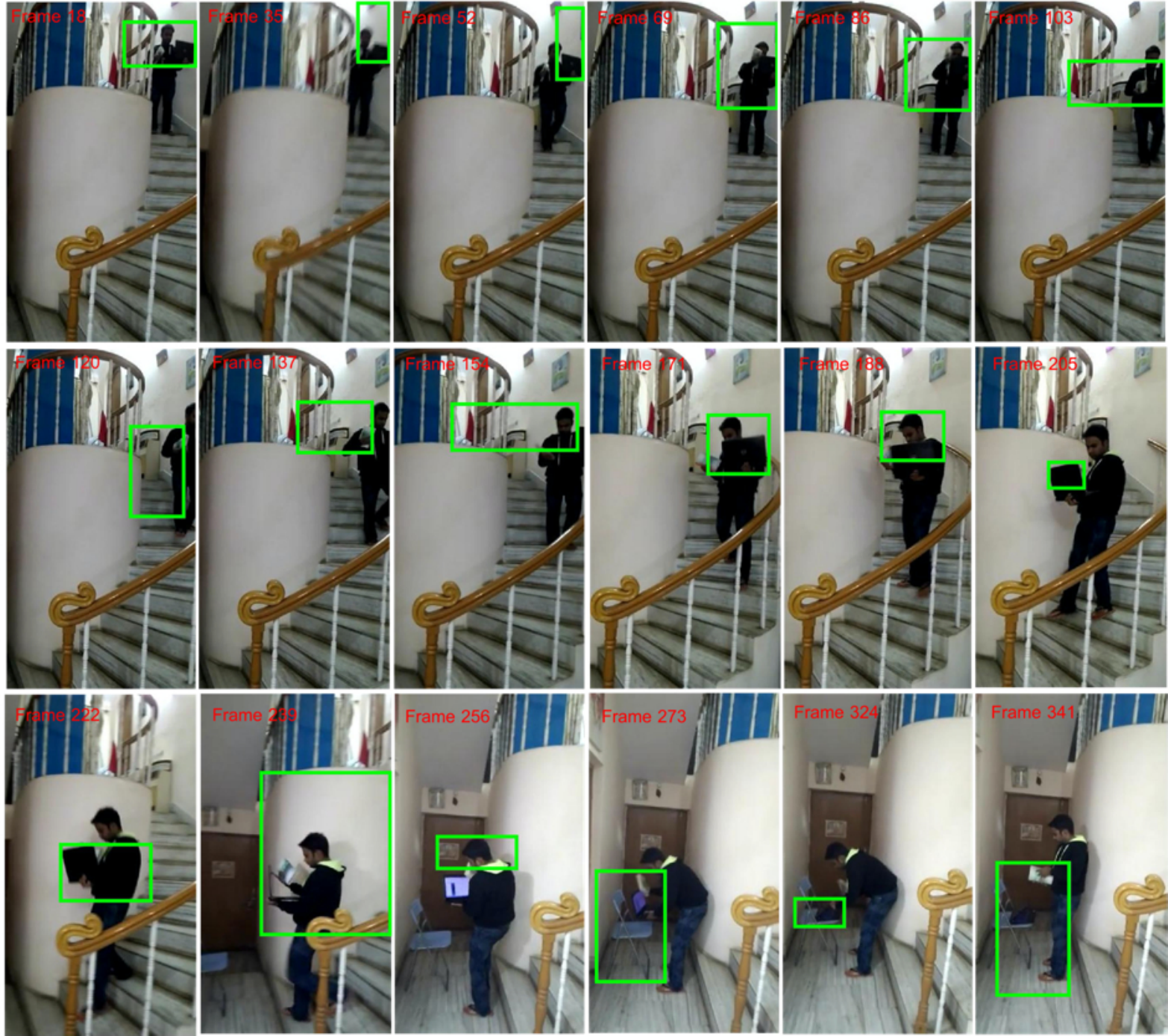}
  		\caption{{Class \textit{Laptop}.}}
  		\label{fig:supp1}
  		\vspace{-6mm}
  	\end{center}
  \end{figure*}
  
  \begin{figure*}[!h]
  	\begin{center}
  		\includegraphics[width=0.9\linewidth]{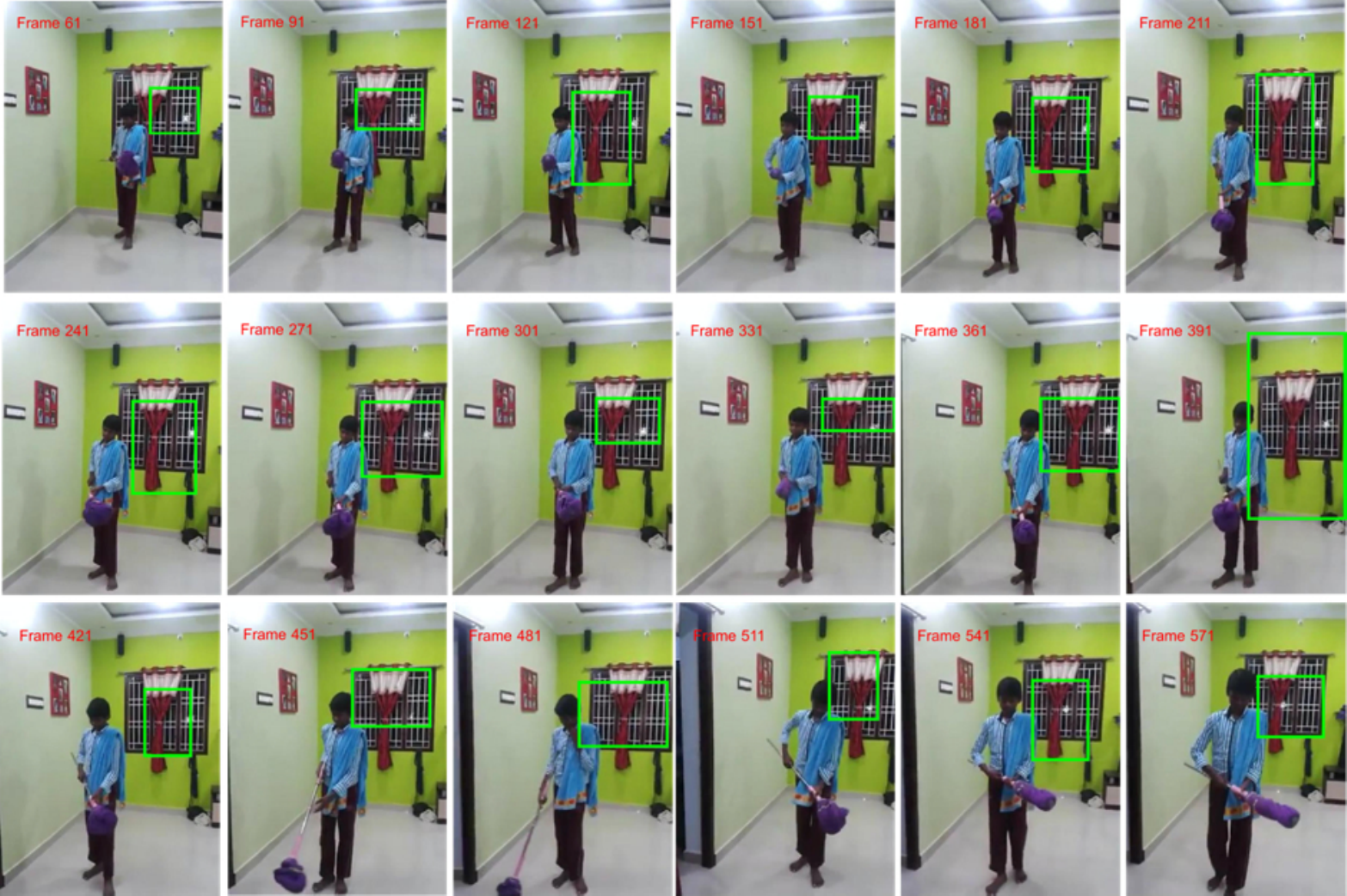}
  		\caption{{Class \textit{window}.}}
  		\label{fig:supp2}
  		\vspace{-6mm}
  	\end{center}
  \end{figure*}
  
  \begin{figure*}[!h]
  	\begin{center}
  		\includegraphics[width=0.9\linewidth]{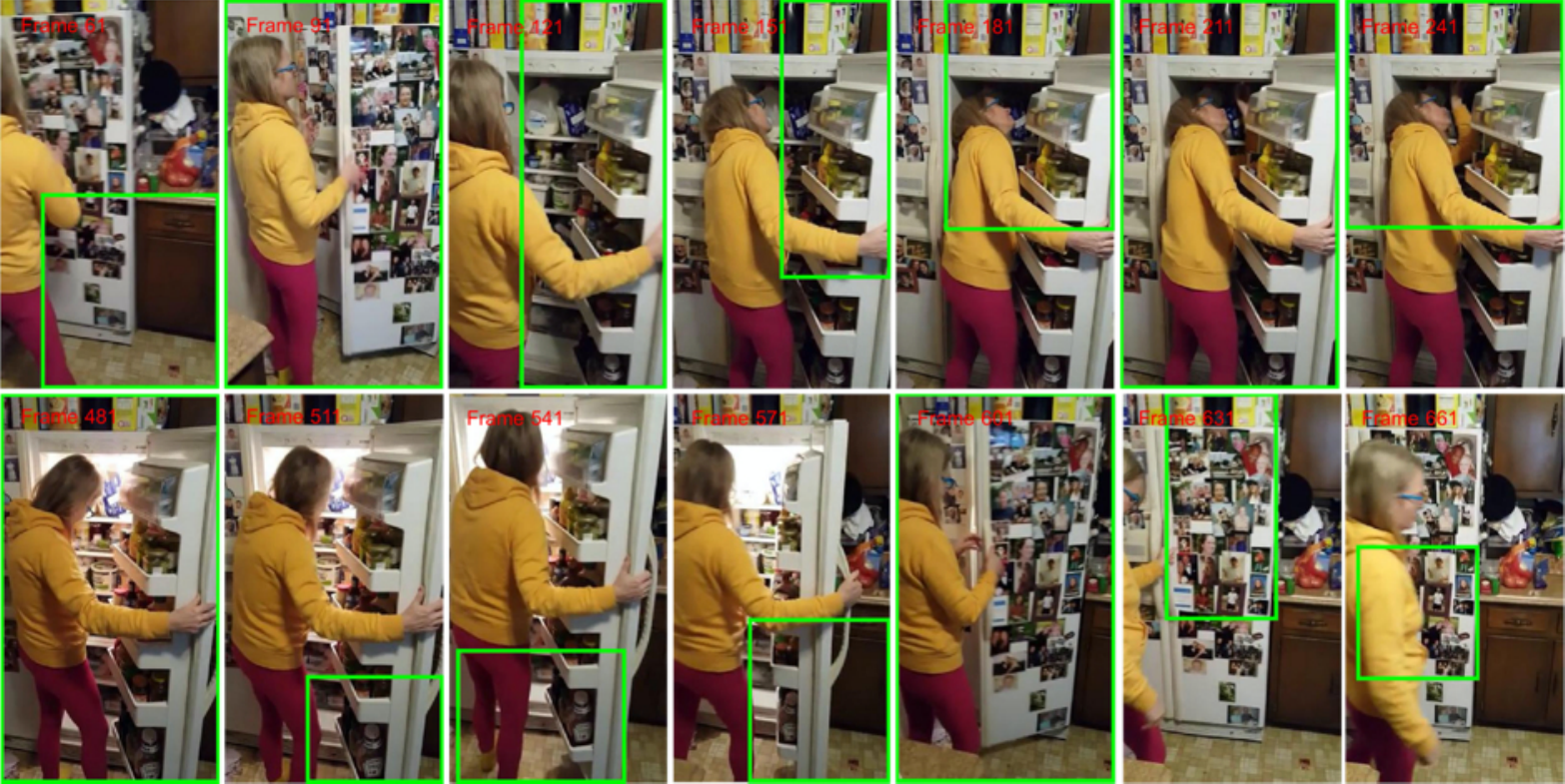}
  		\caption{{Class \textit{Refrigerator}.}}
  		\label{fig:supp3}
  		\vspace{-6mm}
  	\end{center}
  \end{figure*}
  
  \begin{figure*}[!h]
  	\begin{center}
  		\includegraphics[width=0.9\linewidth]{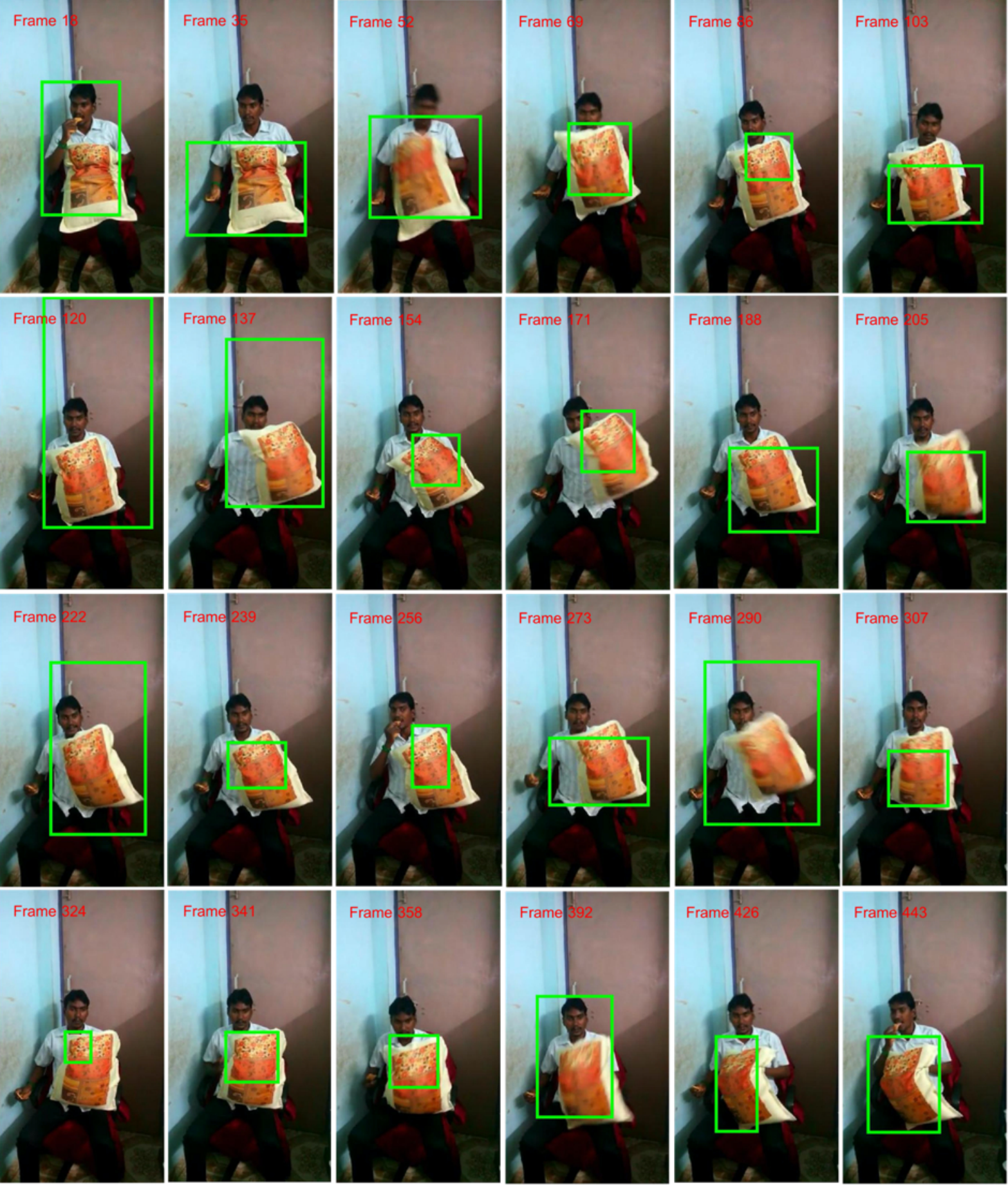}
  		\caption{{Class \textit{Pillow}.}}
  		\label{fig:supp4}
  		\vspace{-6mm}
  	\end{center}
  \end{figure*}

\end{document}